\documentclass[letterpaper, 10 pt, conference]{ieeeconf}  %

\IEEEoverridecommandlockouts                              %

\usepackage{graphics} %
\usepackage{epsfig} %
\usepackage{times} %
\usepackage{amsmath} %
\usepackage{amssymb}  %
\usepackage{siunitx}
\usepackage[dvipsnames,table,xcdraw]{xcolor}
\usepackage{float}
\usepackage{xspace}
\newcommand{\phase}{\phi}
\newcommand{\coupling}{\sigma}
\newcommand{\grf}{F^{\text{GRF}}}
\newcommand{\tonic}{\omega}
\newcommand{\offset}{\xi}
\newcommand{\state}{x}
\newcommand{\kinMap}[1]{g(#1)}
\newcommand{\oscDyn}[1]{f(#1)}
\newcommand{\gaitReward}{r_{\text{gait}}}
\newcommand{\ORC}[1]{ORC(#1)\xspace}
\newcommand{\gaitPhase}{RPD\xspace}
\usepackage{caption}
\usepackage{hyperref}
\usepackage{subcaption}
\usepackage{booktabs}
\usepackage{graphicx}  %
\usepackage{booktabs}  %
\usepackage{multirow}  %
\usepackage{stfloats}

\usepackage[backend=biber,
    style=ieee,
    bibstyle=ieee,
    sortcites=true,   
    minbibnames=10,
    maxbibnames=20,
    maxcitenames=2,
    mincitenames=2,
    giveninits=true,
    uniquelist=false,
    backref=false]{biblatex}
\title{\LARGE \bf
Learning Emergent Gaits with Decentralized Phase Oscillators: \\ on the role of Observations, Rewards, and Feedback
}

\author{Jenny Zhang, Steve Heim, Se Hwan Jeon, Sangbae Kim%
\thanks{All authors are with the Biomimetic Robotics Lab, Massachusetts Institute of Technology, Cambridge, Massachusetts,
            USA (e-mails: \tt\footnotesize
            {jenzhang@alum.mit.edu},
            \tt\footnotesize
            {heim.steve@gmail.com},
            {\tt\footnotesize\{sehwan, sangbae\}@mit.edu})}%
}

\begin{document}

\maketitle

\begin{abstract}
We present a minimal phase oscillator model for learning quadrupedal locomotion. 
Each of the four oscillators is coupled only to itself and its corresponding leg through local feedback of the ground reaction force, which can be interpreted as an observer feedback gain. 
We interpret the oscillator itself as a latent contact state-estimator. 
Through a systematic ablation study, we show that the combination of phase observations, simple phase-based rewards, and the local feedback dynamics induces policies that exhibit emergent gait preferences, while using a reduced set of simple rewards, and without prescribing a specific gait.
The code is open-source, and a video synopsis available at \url{https://youtu.be/1NKQ0rSV3jU}.
\end{abstract}
\section{INTRODUCTION}
Quadrupedal animals exhibit a variety of gaits, or pattern of footfalls, and the choice of gaits has been linked to energetics, speed, morphology, etc.~\cite{hoyt1981gait, wilshin2017longitudinal, hildebrand1968symmetrical}.
Quadrupedal robot controllers, on the other hand, are typically designed around a fixed contact sequence, for a number of reasons.
First, the energetic difference between gaits for robots has been shown to be inconsequential~\cite{bledt2018cheetah, xi2016selecting}.
Perhaps more importantly, using a single gait greatly simplifies the controller design; for conventional model-predictive control, pre-specifying the contact sequence allows a significantly simpler problem to be solved~\cite{kim2019highly, chen2023quadruped, kong2023hybrid}.
Model-free reinforcement learning (RL) side-steps the computational burden of reasoning over different contact sequences.
Nonetheless, RL typically requires extensive reward shaping and regularization, which is often encoded in a time-indexed reference trajectory based on a fixed gait, either as a nominal trajectory~\cite{miki2022learning, feng2023genloco, xie2021dynamics} or a reward~\cite{jin2022high}.
A periodic clock observation is then necessary to maintain the Markov property of the reference.
\par
\begin{figure}[t!]
    \centering
    \includegraphics[width=\columnwidth]{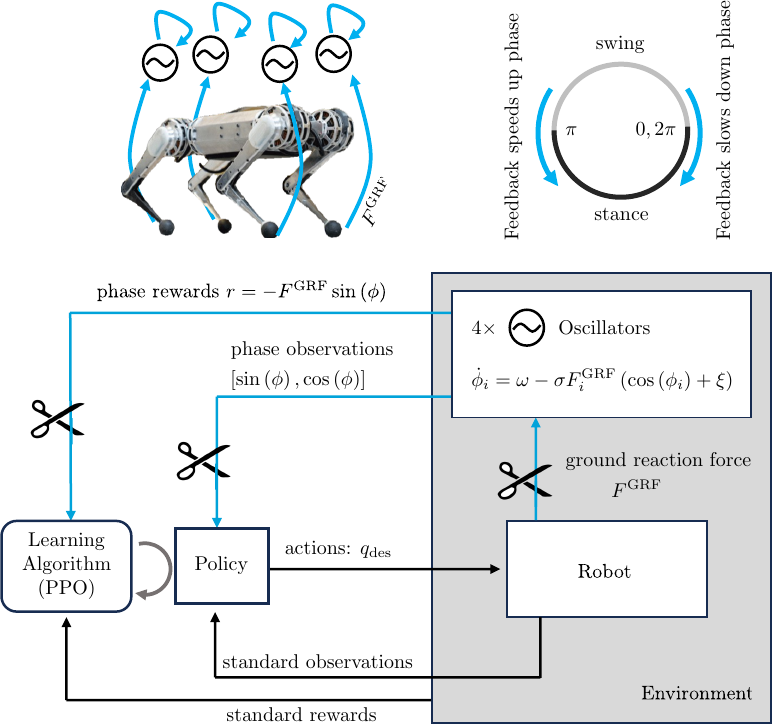}
    \caption{
    We augment the robot state with four decentralized phase oscillators, one per leg. Blue arrows in the diagram indicate the three oscillator-related signals: first, observations of the oscillator phase to the policy to make feed-forward Markov. Second, the phase-based reward encodes the general properties of gaits. Finally, the ground reaction force ($\grf$) is used as feedback, which we view as the observer feedback that allows us to interpret the phase oscillators as a state observer of whether each foot should be in stance or swing. The scissors represent our ablation study. \vspace{-20pt}}
    \label{fig:diagram}
\end{figure}
It is generally difficult to design shaping rewards that capture a general high-level notion, in our context ``locomote with a regular gait'', without over-specifying the solution.
\textcite{siekmann2021allGaits} proposed a simple phase-based reward to encourage stance or swing at any phase difference for their bipedal robot Cassie, and demonstrated this policy can then track any desired gait by simply adjusting the phase difference between the legs accordingly.
Similar work has applied this type of reward to quadrupeds as well~\cite{margolis2023theseWays, shao2021learning}.
In these cases, the actual gait choice is removed from the policy, which is essentially treated as a low-level policy.
Instead, the phase difference needs to be specified by the user or a high-level policy~\cite{yang2022fast}.
\par
A popular approach to achieving specific phase differences is to augment the state space with a network of coupled phase oscillators called central pattern generators (CPGs)~\cite{ijspeert2008central}.
In its simplest form, the dynamics of the oscillators are designed to exhibit stable limit cycles, and the resulting phases are mapped to values relevant to the robot controller, such as desired kinematics.
In essence, this is a generalization of time-indexed trajectories.
Despite the often simpler and lower-dimensional design space of the phase oscillators (compared to the space of reference trajectories in joint space), it can still be difficult to design the oscillator dynamics and mapping~\cite{buchli2006engineering}.
Recent works have instead opted to learn the coupling, while fixing the mapping from the phase to robot states~\cite{bellegarda2022cpg, yang2022fast}.
This approach complements the phase-based low-level policies, and CPGs are often interpreted as feed-forward reference generators~\cite{ijspeert2023integration, raffin2024simpeBasline}.
An alternative view, recently proposed by~\textcite{ryu2021optimality}, interprets the CPG as an observer, with reflex-like sensory feedback playing the role of the observer gain.
\par
We present an implementation of \emph{decentralized} phase oscillators based on the work of~\textcite{owaki2013simple}, and, based on the observer-interpretation of~\textcite{ryu2021optimality}, treat the oscillator phase as a loose estimate of whether each leg should be in swing or stance.
Based on this interpretation, we use a reward similar to~\textcite{siekmann2021allGaits} to encourage the policy and phase oscillators to entrain.
We further enrich the oscillator dynamics with an offset term to entrain the phases for standing still while retaining the same reward structure.
\par
The three key signals afforded by our architecture, represented in Fig.~\ref{fig:diagram} with blue arrows, are the \emph{phase observation} which renders feed-forward policies Markov, the \emph{phase rewards} that encode the high-level gait properties of duty factor and nominal frequency, and the \emph{local feedback coupling} through the ground reaction force coupling the oscillator dynamics and the policy.
We will show the importance of each of these through a systematic ablation.
\section{PRELIMINARIES}\label{sec:prelim}
We introduce gaits and pattern generation based on phase oscillators. We assume working knowledge of reinforcement learning and quadrupedal robots; for the interested reader, we recommend~\cite{rudin2022walkMinutes, kober2013reinforcement}.

\subsection{Gaits}\label{subsec:gaits}

For the scope of this paper, we restrict ourselves to quadrupedal gaits, which are defined by the pattern of stance phases~\cite{hildebrand1968symmetrical}.
In the rest of this paper, we will define gaits by the relative phase difference (\gaitPhase); using the right front (RF) foot as the reference, we first define the gait cycle length as the time between consecutive RF foot touchdowns, normalized to $2\pi$.
We then calculate the \gaitPhase as a 3D vector, composed of the time difference between touchdowns of the left front (LF), right hind (RH), and left hind (LH) feet to the RF reference foot, normalized by the gait cycle length.
Each gait is fully defined by these three values. The ideal symmetric gaits
\textbf{Trot} $(\pi, \pi, 0)$,
\textbf{Pace} $(\pi, 0, \pi)$,
\textbf{Bound} $(0, \pi, \pi)$,
\textbf{Pronk} $(0, 0, 0)$
are marked in Fig.~\ref{fig:rpdPlots}.
\par
We classify gaits by averaging the \gaitPhase over two gait cycles every 5 seconds, then taking the closest ideal gait within the set of symmetric quadruped gaits by Euclidean distance in RPD space.
If the distance to the closest ideal gait exceeds 2, we classify the \gaitPhase as being in \textbf{Transition}.

\subsection{Phase-based Pattern Generation}\label{subsec:patternGeneration}
We will distinguish between clocks, central pattern generators (CPGs) without feedback, CPGs with feedback, and decentralized oscillators (which are driven by feedback by definition).
Each of these is a special case of a system of oscillators with state $\phase \in \left[0, 2\pi\right)^n$, where $n \in \mathbb{N}$ is the number of oscillators, and
\begin{equation}\label{eq:general_oscillator}
\begin{split}
\dot{\phase}_i &= \tonic + \oscDyn{\phase_j, \state_j}, \  j \in C(i) \\
    u &= \kinMap{\phase_j, \state_j} \\
    \text{for } i &\in \{1, \ldots, n\}, \text{ and } C(i) \subseteq \{1, \ldots, n\} 
\end{split}
\end{equation}
where $\tonic$ is a nominal frequency, $\state$ is the system state, $\oscDyn{}$ is a function that determines the dynamics properties,
$\kinMap{}$ is a mapping function to some control-relevant input $u$, and the set $C()$ indicates which oscillators are directly coupled.
We will generally consider the case where there is one oscillator per leg, that is $n=4$ for a quadruped.
\\
\textbf{Clocks: $\oscDyn{} = 0$} \\
This degenerate choice for $\oscDyn{}$ reduces the oscillator to a clock with a constant growth rate $\tonic$.
Though sometimes called CPGs~\cite{miki2022learning}, we distinguish this setting as it makes no use of the state and dynamics of the oscillator: the burden is placed on designing the map $\kinMap{}$.
This is the minimal form needed to make a cyclic feed-forward pattern Markov.
\textcite{siekmann2021allGaits} use this setup, and learn the mapping function $\kinMap{\phase, \state_j}$, using the clock as both an observation and to design a simple reward function.
\\
\textbf{CPGs without feedback: $\oscDyn{} = \oscDyn{\phase_j}$}
\\
This form provides a pure feed-forward pattern, and allows the engineer to design the oscillator dynamics, such as limit cycles and convergence properties~\cite{buchli2006engineering, ajallooeian2013general}, unencumbered by the physical dynamics of the robot.
This can significantly simplify the design of a high-level controller that switches between multiple $\oscDyn{}$~\cite{ijspeert2007swimming}: the engineer can ensure smooth transitions by simply enforcing the desired properties in the phase oscillator space, and given a smooth mapping $\kinMap{\phi}$, those properties are retained in the generated reference trajectory.
However, once the phase has converged to the limit cycle, this setup effectively acts as a clock, as the phase oscillator state cannot be perturbed off the limit cycle.
\\
\textbf{CPGs with feedback: $\oscDyn{} = \oscDyn{\phase_j, \state_j}$}
\\
Coupling the phase oscillator and physical states fully exploits the dynamics properties of the phase oscillator, but also makes it more difficult to design useful dynamics.
Nonetheless, even relatively simple, local feedback has been shown to greatly improve performance~\cite{ajallooeian2013modular, aoi2013stability}.
Several studies have relied on RL to learn $\oscDyn{}$~\cite{bellegarda2022cpg, yang2022fast}.
\textcite{ryu2021optimality} also use this form, but re-interpret it as a state estimator, with the state feedback acting as the observer gain.
We rely heavily on this interpretation.
\\
\textbf{Decentralized Oscillators:} $C(i) = \left\{i\right\}$ \\ %
This is a special case where $\oscDyn{} = \oscDyn{\phase_i, \state_i}$: each oscillator is only affected by feedback from itself and sensory information from its corresponding leg.
Despite lacking direct coupling between oscillators, such systems can still synchronize due to the indirect feedback coupling through local physical interaction~\cite{owaki2013simple, schilling2020decentralized}.
Our work uses this setup, and in particular a form based on the so-called \emph{Tegotae} feedback model proposed by~\textcite{owaki2013simple}:
\begin{equation}\label{eq:tegotae}
    \oscDyn{} = -\coupling \grf_i \left (\cos\left(\phase_i\right) \right )
\end{equation}
where $\coupling$ is a feedback gain, and $\grf_i$ is the ground reaction force felt at foot $i$.
\par
This model captures the essence of a gait (frequency, swing/stance phases) without specifying any details.
Indeed, it would be difficult to predict any specific gait when looking at Eq.~\eqref{eq:tegotae}.
This type of decentralized oscillators has been shown to not only converge to gaits, but also switch gaits depending on forward velocity~\cite{owaki2013simple, owaki2017quadruped}.
\par
We use a variation of this oscillator with the interpretation proposed by~\textcite{ryu2021optimality}, that each phase is a latent observation of whether the corresponding leg should be in stance or swing, and $\coupling$ in Eq.~\eqref{eq:tegotae} is an observer gain.
\section{Implementation details}
We train policies and collect simulation data in IsaacGym, using a fork of \texttt{legged\_gym} by~\textcite{rudin2022walkMinutes}, and proximal policy optimization.
Our code is open-sourced~\cite{ORCAgym} and can be referenced for implementation details, hyperparameters, and instructions to reproduce our results.
\subsection{Robot}
We use the MIT Mini Cheetah robot~\cite{katz2019mini} in simulation and for hardware transfer.
For all results presented in the paper, which rely on large-scale data collection for statistical analysis, we use simulation results.
Preliminary tests on hardware can be seen in the supplementary video.
\\\textbf{Observation space:} The policy is given standard state observations composed of the base angular velocity, projected gravity vector, joint positions and velocities, the previous actions, and the desired base linear and angular velocity commands.
When not ablated, the observations also include the phase oscillator observations $\left[\sin\left(\phase_i\right), \cos\left(\phase_i\right)\right], i \in \{\text{FR}, \text{FL}, \text{HR}, \text{HL}\}$.
The critic is additionally given the current oscillator velocity as privileged information.
\\
\textbf{Action space:} The policy outputs desired joint positions at 100 Hz, which are fed into a low-gain PD-controller with $K_p = 20$ and $K_d = 0.5$ running at 500 Hz.
\\
\textbf{Rewards:} Following standard conventions, we use positive rewards for the command tracking error, orientation error, and a minimum base-height error, all passed through a squared exponential function.
In addition, we regularize with negative rewards on the square of the torques, first and second-order action smoothness, and the hip abduction/adduction joints deviating from the resting position.
Finally, body collisions with the ground terminate the episode and incur a flat penalty.
\par

\par
When not ablated, we also add the phase-based reward
\begin{equation}
    \gaitReward(\state) = - \grf_i \sin\left(\phase_i\right)
\end{equation}
which penalizes foot contact when $\phase_i \in \left[0, \pi\right)$ and encourages contact during $\phase_i \in \left[\pi, 2\pi\right)$.
Due to the dynamics of the phase oscillator, this simple reward encourages two effects: first, to learn a policy with roughly the oscillator nominal frequency $\tonic$, and second, to have consistent periods of contact with all feet.
We will see that the policy also learns to actively use the $\grf$ to `guide' the oscillators into stable gaits if the coupling is included during training.
\\
\textbf{Ground Reaction Force Estimation:} Each $\grf_i$ value is normalized by robot mass and clamped between $\left[0, 1\right]$, such that $\coupling$ values can be kept consistent for robots of different sizes.
We also found this mitigated issues caused by the highly inaccurate contact force estimates in IsaacGym.

\subsection{Decentralized phase oscillators}\label{subsec:oscImplementation}
We add an offset term $\offset$ to the original \emph{Tegotae} feedback model from Eq.~\eqref{eq:tegotae}, so our oscillator dynamics become
\begin{equation}\label{eq:ourOscillator}
    \dot{\phase}_i = 2\pi\left(\tonic -\coupling \grf_i \left (\cos\left(\phase_i\right) + \offset \right )\right)
\end{equation}
Based on biological observations~\cite{heglund1974scaling} and trial and error, the parameters $\tonic$, $\coupling$, and $\offset$ are set to
\begin{equation}\label{eq:ramp}
    [\tonic,\, \coupling,\, \offset](v_x)= 
\begin{cases}
    [1,\, 4,\, 1] & \text{if } |v_x|\leq 0.5\\
    [\min\{1.5+|v_x|, 4\},\, 1,\, 0] & \text{otherwise}
\end{cases}
\end{equation}
where $v_x$ is the commanded forward velocity.
Each phase $\phase_i$ is uniformly randomized at the start of each episode.
\par
\begin{figure}[htbp]
\centerline{\includegraphics[width=\columnwidth]{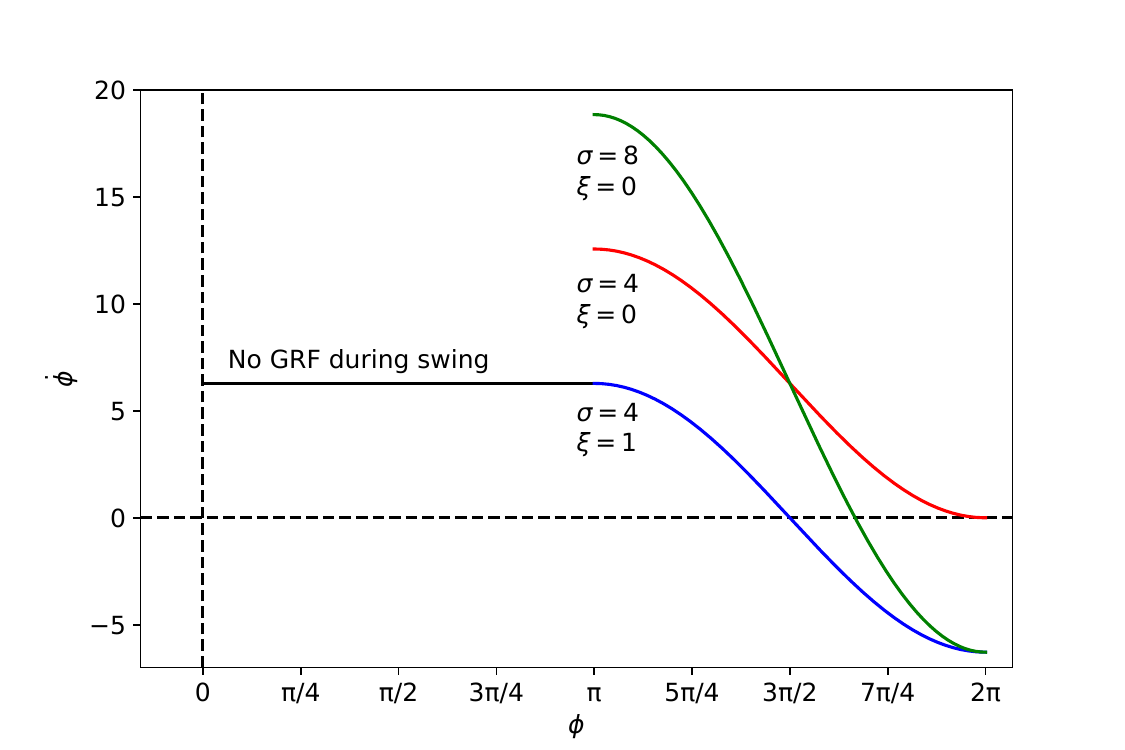}}
\caption{
To illustrate the oscillator fixed points (where $\dot{\phi}=0$), we graph the oscillator dynamics with $\grf=0.25$, where each leg supports a quarter of the body weight.
For the curve with coupling $\coupling=4$ and offset $\offset=0$, the point at phase $\phase=2\pi$ is only marginally stable, and would not settle in stance.
When $\offset=0$, the limit of the fixed point as $\coupling$ approaches $+\infty$ is $3\pi/2$, but drastically increasing $\coupling$ alone introduces discrete jumps in $\phase$ that are destabilizing. Setting $\offset=1$ with $\coupling=4$ caps $\dot{\phase}$ at the nominal $2\pi\tonic$, and places the fixed point directly in the middle of the stance phase.
This formulation helps to achieve standing without stepping in place. \vspace{-12pt}}
\label{fig:offset}
\end{figure}

Standing still is a special case for which we want all oscillators to settle to a stable point in stance.
Rather than activating special rewards to minimize joint velocities when the commanded velocity is low, we change the oscillator parameters to smoothly introduce a stable fixed point.
This allows the fixed point to be closer to the middle of the stance range $\left[\pi, 2\pi\right)$ without using an excessively large $\coupling$ value, which can destabilize the oscillator by chattering between stance and swing phases.
This smoothing effect is illustrated in Fig.~\ref{fig:offset}, with the blue line showing the actual phase velocity vs. phase function used during training for zero velocity commands.
\vspace{-20pt}
\section{Observation, Reward, Coupling Ablation}\label{sec:results}
We perform an ablation of the phase-based observations, rewards, and coupling by cutting their respective signals during training, see Fig.~\ref{fig:diagram}.
We will refer to each permutation as \ORC{xxx}, where each entry is a boolean indicating whether the corresponding signal is present or not during training. For example, \ORC{110} indicates a policy that was trained with the phase observations and phase-based rewards, but with the coupling term $\coupling = 0$. Note that \ORC{110} is essentially the setting used by~\textcite{siekmann2021allGaits}.
\par
We skip the permutation \ORC{001} as it is equivalent to \ORC{000}.
For all other permutations, we train and statistically analyze 10 policies each to answer three questions: \\
\ref{subsec:balance}) \emph{How do the signals influence leg-load distribution?} \\
\ref{subsec:gaitemergence}) \emph{How do the signals influence gait emergence?} \\
\ref{subsec:disturbance}) \emph{How does each signal affect overall stability?}
\subsection{Balanced Leg Use}\label{subsec:balance}
Ideally, a policy should use all four legs in a consistent and balanced manner.
\ORC{x0x} policies either don't have oscillator observations or are not encouraged to use them in any particular way.
\ORC{01x} policies are non-Markov as it appears to receive inconsistent rewards for performing the same action given its observable state.
\ORC{11x} policies are trained with both oscillator observations and rewards to encourage swing and stance during specific ranges of the oscillators, and have enough information to match phases with rewards related to ground contact.
We expect to see more consistent leg use with the policies trained using \ORC{11x}, and more variability in the other policies.
\par
\begin{figure}[htbp]
\vspace{-10pt}
\centerline{\includegraphics[width=\columnwidth]{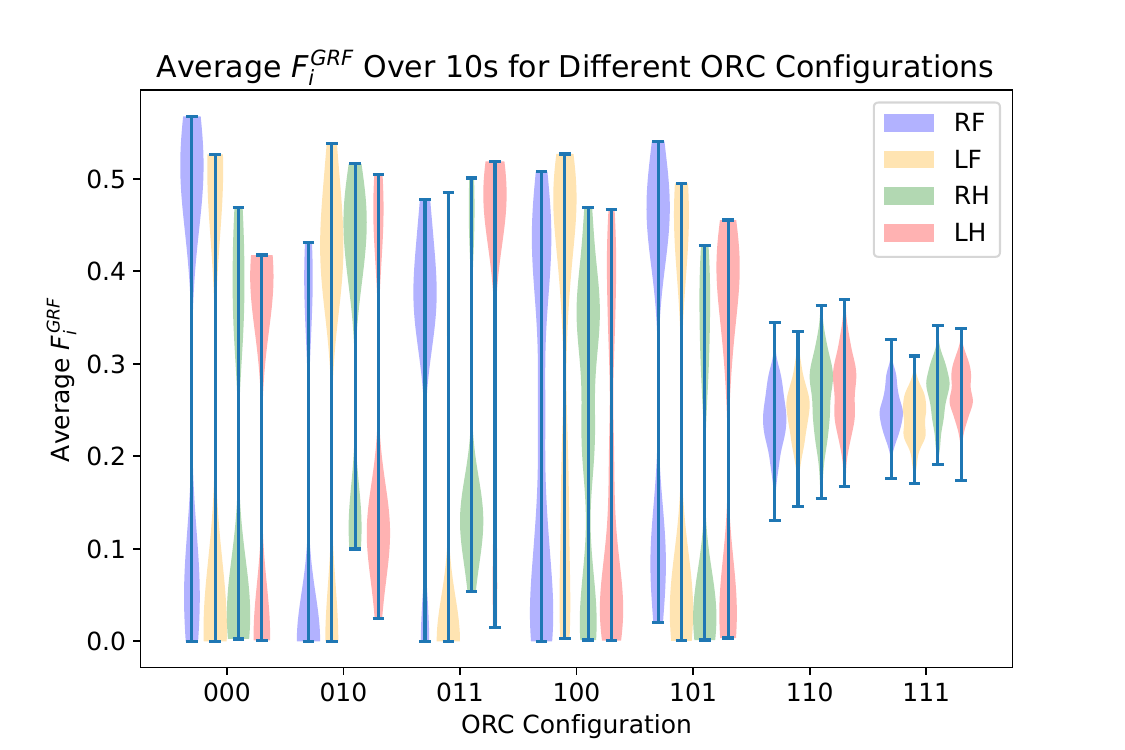}}
\caption{Each ORC configuration has 500 agents (50 per re-trained policy), and $\grf$ is averaged for each leg across the entire episode. \ORC{11x} policies show much more consistent and balanced leg use compared to all other configurations, which tend to exhibit 2 or 3 legged gaits. \vspace{-5pt}}
\label{fig:legvariance}
\end{figure}

For each policy, 50 robots are initialized with random oscillator phases and rolled out in simulation with 1 m/s forward velocity command.
We calculate the average $\grf_i$ over the entire 10 second episode for each leg separately.
Randomly initialized oscillator phases result in different behavior in the 50 agents for \ORC{1xx} policies.
Results from all 500 runs belonging to each ORC configuration are aggregated into the same dataset.
Each violin plot in Fig.~\ref{fig:legvariance} shows the distribution of average $\grf_i$ experienced by each leg.
Since $\grf_i$ is normalized by body weight, perfectly balanced leg use yields average $\grf_i = 0.25$ for all legs.

\par
We can see that for all ORC configurations without observations and rewards working together, the distributions of $\grf_i$ are very wide and have significant clusters around 0 for at least one leg out of four.
This agrees with our observation that the policies trained with those ORC configurations often result in 2 or 3 legged gaits, where some feet are dragging or always kept in the air, and therefore experiencing little to no $\grf_i$ over the roll-out.
The pattern of two clusters around the extremities of the distributions for each leg arises from inconsistent policies after repeated training with the same setting.
In some policies, the RF foot might always be in the air, while in others the LF foot might always be in the air.
\par
The phase observations and rewards strongly encourage \ORC{11x} policies to use all four legs cyclically without specifying the exact desired gait.
Both configurations yield distributions that are roughly centered around average $\grf_i = 0.25$, showing that the leg use is well balanced over all trials and policies.
\par
\ORC{110} experiments have slightly larger range compared to \ORC{111}, which may be attributed to some randomly initialized asymmetric gaits requiring higher $\grf_i$ on some legs compared to the others to track, since it cannot converge to a more symmetric gait without coupling.

\subsection{Emergence of Gaits}\label{subsec:gaitemergence}
To evaluate gait emergence, we focus only on \ORC{110} and \ORC{111} policies, since we see from {Subsection~\ref{subsec:balance}} that other permutations rarely yield well-defined gaits.
Each experiment evaluates a single policy chosen at random, and includes 500 runs with randomized initial oscillator values, rolled out over 40 seconds with a 1 m/s forward velocity command.
We calculate the \gaitPhase as described in Subsection~\ref{subsec:gaits}, and assess gait preference by evaluating the \gaitPhase distribution at the end of the roll-out.
We verify that \ORC{111} policies don't ignore the phase observation by deactivating the coupling at execution time.
We further probe the role of the feedback coupling by activating it for an \ORC{110} policy, which was trained with no coupling.
\par
\begin{figure*}[t]
\vspace{10pt}
\centerline{\includegraphics[width=2\columnwidth]{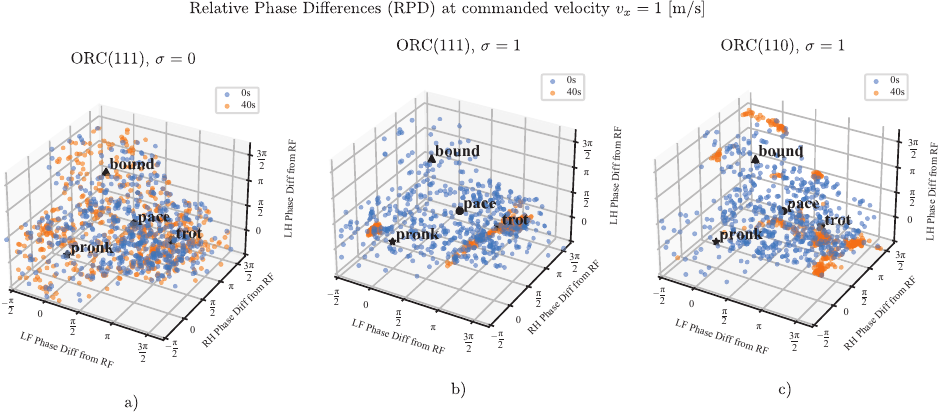}}
\caption{Initial and final relative phase difference \gaitPhase points are shown for 500 randomly initialized runs in each experiment. \ORC{111} evaluated with coupling $\coupling=0$ tracks the initial phases and cannot converge to any specific gait. \ORC{111} evaluated with $\coupling=1$ exhibits strong convergence to trot and pronk, while \ORC{110} evaluated with $\coupling=1$ exhibits some convergence around trot, but is more spread out compared to the final \ORC{111} \gaitPhase.} \vspace{-15pt}
\label{fig:rpdPlots}
\end{figure*}

Both \ORC{111}, visualized in Fig.~\ref{fig:rpdPlots}a), and \ORC{110} policies match their \gaitPhase to the oscillator phases when the coupling is set to $\coupling=0$, as expected.
\par
When evaluating \ORC{111} with coupling $\coupling=1$ (same as during training), the gaits converge within 10 seconds to their final preferred state, and we observe mostly trotting and pronking gaits in tight clusters on Fig.~\ref{fig:rpdPlots}b).
Since the policy's actions influences the oscillators through ground contacts, it can learn to manipulate the randomly initialized oscillators and phase lock into desirable gaits faster.

\begin{figure}[htbp]
\centerline{\includegraphics[width=\columnwidth]{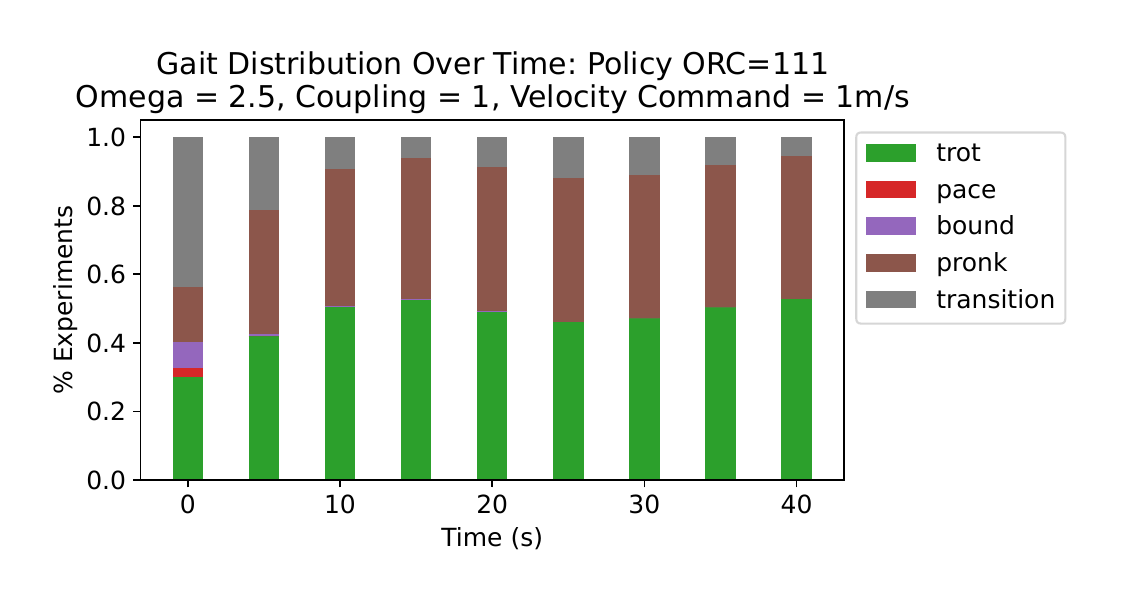}}
\caption{The distribution of gaits for 500 runs of \ORC{111} with $\coupling=1$ settles into both trot and bound quickly within 10 seconds.} \vspace{-15pt}
\label{fig:gait111eval111}
\end{figure}
\begin{figure}[htbp]
\centerline{\includegraphics[width=\columnwidth]{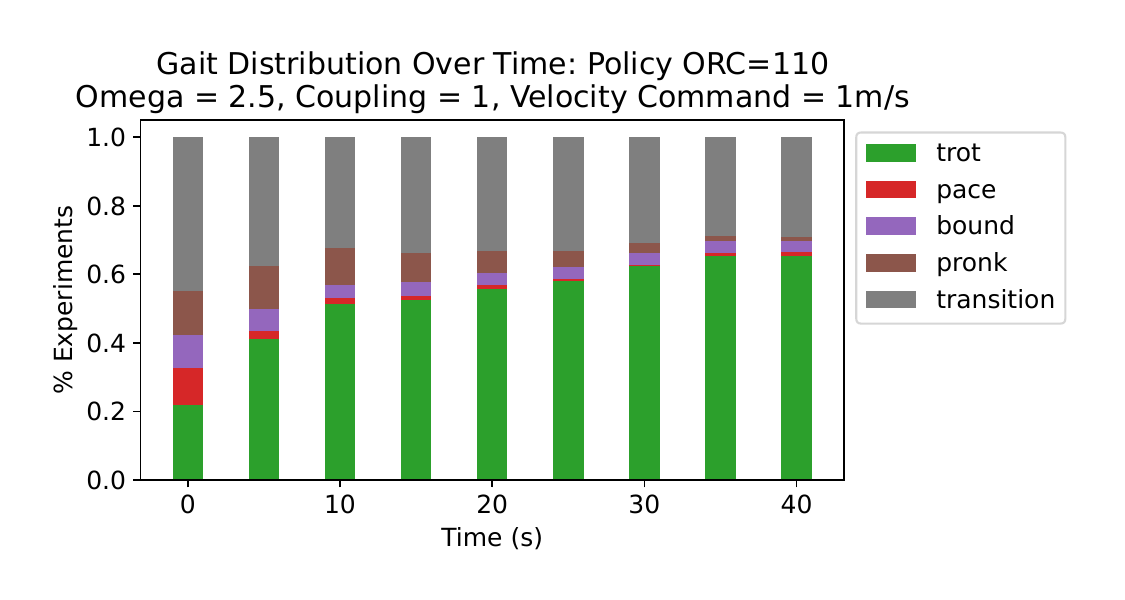}}
\caption{The distribution of gaits for 500 runs of \ORC{110} with coupling $\coupling=1$ settles into trot slowly, with more environments transitioning at the beginning but others continuing to slowly converge toward trot as runs with \gaitPhase initialized further away become more trot-like over time due to the oscillator dynamics.} \vspace{-20pt}
\label{fig:gait110eval111}
\end{figure}

When evaluating \ORC{110} with $\coupling=1$, the policy continues to track the oscillators, which are now also modulated by the feedback term that was not seen during training. 
The gaits that emerge from this experiment have regions of attraction dictated by the oscillator dynamics, which highlights the role of the oscillator in determining the preferred gait.
However, as shown in Fig.~\ref{fig:rpdPlots}c), the final gaits are clustered further away from the ideal phase differences of symmetric gaits compared to the \ORC{111} policy shown in Fig.~\ref{fig:rpdPlots}b).
We also see in Fig.~\ref{fig:gait111eval111} and Fig.~\ref{fig:gait110eval111} that the \ORC{110} (evaluated with coupling) takes nearly 40 seconds to settle into its preferred gait, whereas the \ORC{111} policy settles in roughly 10 seconds.

\par
\begin{figure}[htbp]
\centerline{\includegraphics[width=\columnwidth]{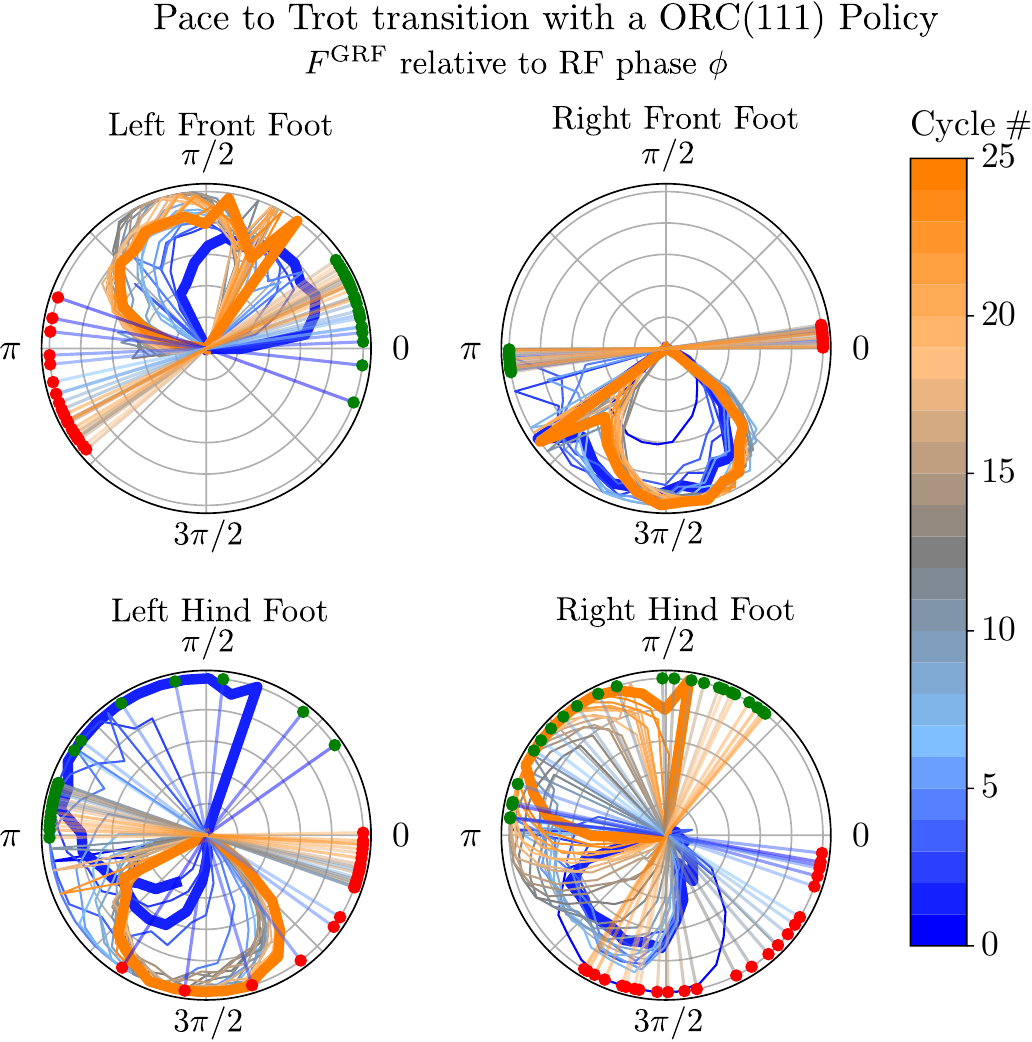}}
\caption{The ground reaction force $\grf_i$ is plotted for each foot relative to $\phase_{\text{RF}}$, with initial gait cycles in blue and progressing through time to orange.
The phase $0$ and $\pi$ crossings of each leg are shown with red and green dots, respectively.
The bold cycles are the initial and final cycles, which show this run starting in pace with lateral feet in phase and ending in trot with diagonal feet in phase.
All swing and stance behavior obeys each leg's respective oscillator phase well, with non-zero $\grf$ falling between the green dot $\pi$ crossings and red dot $0$ crossings in every gait cycle. \vspace{-15pt}}
\label{fig:pacetotrot}
\end{figure}

Fig.~\ref{fig:pacetotrot} shows a pace to trot transition of one of the runs from Fig.~\ref{fig:gait111eval111} by plotting all $\grf_i$ from 0-10 seconds with respect to the oscillator phases of the RF reference leg.
The thick blue lines show $\grf_i$ during the first gait cycle and the thick orange lines show $\grf_i$ during the last gait cycle.
As more gait cycles occur, the phase differences between the leg oscillators change, indicated by the red and green dots showing the RF phase value at the time-step when the corresponding leg's oscillator crosses 0 and $\pi$ respectively.
Those dots do not exactly overlap for the RF leg because of discrete time-step errors.
The $\grf_i$ of each leg follows its own phases, and over time settles into a trotting gait, with diagonal legs being in phase.

\par

\subsection{Disturbance Rejection}\label{subsec:disturbance}
To compare the overall effect on stability, we test how well each policy can reject a planar velocity impulse applied to the body while being commanded to run at 3 \si{m/s}, with oscillators being initialized at random and allowed a 5 second settling period.
To ensure that we apply perturbations at all phases of the gait, we run 1800 trials per policy, and stagger the perturbations: every 0.01 seconds, a ball of perturbations spaced at 10 degree intervals is applied to a different set of 36 robots, over a period of 0.5 seconds.
As all policies exhibit a frequency of roughly 4 Hz at this commanded speed\footnote{We verified this frequency via a fast Fourier transform on the joint positions for policies without phase observations.}, staggering the perturbations ensures that we apply perturbations at different phases of the gait, with different numbers of feet on the ground.
We then calculate the mean and standard deviation of the failure rate across all ten policies learned for each permutation of \ORC{xxx}.
\par
The failure rate means and standard deviations, reported in Table~\ref{tab:pushball}, show that policies trained under \ORC{111} consistently have a lower failure rate compared to all other permutations.
Surprisingly, toggling on phase-based rewards has a stronger impact than the phase-based observations, even for cases such as \ORC{010} and \ORC{011} where the reward is not Markov.
We conjecture that, when the reward is not Markov, it should be viewed as a stochastic reward that still discourages chattering contacts or dragging feet, despite not signaling any specific gait schedule.
\begin{table*}[tbh]
\vspace{5px}
\caption{Failure rate after velocity impulse disturbance (All values in \%)}\label{tab:pushball}
\centering
\begin{tabular}{c|l|ccccccc}
   ORC & & 000 & 100 & 101 & 010 & 110 & 011 & 111 \\
\midrule
\multirow{7}{*}{\rotatebox{90}{\parbox{0.8cm}{Impulse \\ magnitude [\si{m/s}]}}}
& 1.5 &    \textbf{0.0  \(\pm\) 0.0}  &    \textbf{0.0  \(\pm\) 0.0}  &    0.1  \(\pm\) 0.2  &   \textbf{0.0  \(\pm\) 0.0}  &    \textbf{0.0  \(\pm\) 0.0}  &    \textbf{0.0  \(\pm\) 0.0}  &    \textbf{0.0  \(\pm\) 0.0} \\
& 2.0 &     1.7 \(\pm\) 2.5  &     4.1 \(\pm\) 5.2  &    0.9 \(\pm\) 1.2  &    \textbf{0.0  \(\pm\) 0.0}  &    0.0 \(\pm\) 0.1  &    \textbf{0.0 \(\pm\) 0.0}  &    0.0 \(\pm\) 0.1 \\
& 2.5 &    11.0 \(\pm\) 13.0  &   10.6 \(\pm\) 9.1  &     7.2 \(\pm\) 6.3  &    0.7 \(\pm\) 0.7  &    0.7 \(\pm\) 1.0  &    \textbf{0.3 \(\pm\) 0.4}  &    0.4 \(\pm\) 0.4  \\
& 3.0 &    16.3 \(\pm\) 12.4  &    17.5 \(\pm\) 8.9  &    16.0 \(\pm\) 10.2  &    3.2 \(\pm\) 3.0   &     4.1 \(\pm\) 3.6  &    3.2  \(\pm\) 3.1  &    \textbf{2.2 \(\pm\) 1.7} \\
& 3.5 &    25.0 \(\pm\) 13.5  &    29.7 \(\pm\) 12.7  &    24.0 \(\pm\) 17.1  &    10.6 \(\pm\) 8.0 &    13.8 \(\pm\) 7.8   &    13.1 \(\pm\) 9.2  &    \textbf{7.4 \(\pm\) 4.0}   \\
\bottomrule
\end{tabular}
\end{table*}
\section{Discussion}
We presented an augmentation of the quadruped robot state space using one decentralized phase oscillator per leg, with a simple feedback coupling to the ground reaction force of the corresponding leg, which can be interpreted as an observer gain.
Through a systematic ablation study, we investigated the importance of each phase-related signal: observations of the oscillator phase, phase-based rewards to encourage distinct swing and stance phases, and feedback coupling.
\par
Overall, \ORC{111} policies trained with all three signals demonstrated the fastest convergence to well-defined gaits, and were consistently the most robust to large impulse perturbations.
We did not find significant differences in local stability\footnote{Analysis of Floquet multipliers and rate of entropy decay were evaluated.} between the policies, which matches our experience in hardware that local stability is not a useful proxy for legged system `stability'.
Nonetheless, preliminary sim-to-real trials (see supplementary video) show significantly better transfer with the observation signal.
~\textcite{bellegarda2022cpg} also reported more reliable sim-to-real transfer when learning a CPG with feedback, although in their case the phase is directly mapped to desired kinematics with a pre-designed mapping.
\par
\ORC{110} policies trained with phase observations and rewards but no coupling showed only slightly worse performance to those with coupling, while tracking the gait defined by whichever phase-difference the oscillators are initialized in, similar to the results of~\textcite{siekmann2021allGaits}.
However, although activating the coupling during evaluation does cause these policies to converge toward symmetric gaits, the convergence time is nearly four times slower.
This observation suggests that the dual roles of control and estimation are not fully separated between the oscillators and the policy, as it is in the linearized model presented by~\textcite{ryu2021optimality}: the policy is affected by the oscillator state, but can also learn to actively drive it towards a more stable gait if trained with the coupling active.
This also matches the observation of~\textcite{ijspeert2023integration} that CPGs may act as both an observer and a pattern generator.
\par
Surprisingly, we found that the reward signal has a stronger effect on stability than the phase observations, despite being non-Markov in some ablations.
Nonetheless, only when both observation and reward signals were present during training, did policies consistently train to exhibit gaits with balanced load distribution among the legs.
\par
Anecdotally, before we introduced the offset term in equation~\eqref{eq:ourOscillator}, policies did not settle into standing as well, but did appear to favor gaits other than pronking more compared to the results presented.
We conjecture that frequent standing causes the oscillators to all sync to stance, and thus biases training data towards pronking gaits.
We also observed different gaits to emerge more frequently at different velocities, or with different morphologies.
Although quantifying the exact convergence and transition patterns is out of the scope of this paper, future work studying the effects of changing oscillator parameters along with physical parameters such as center of mass location and leg length on gait emergence for a quadruped robot could yield connections to gait patterns observed in nature for animals of different sizes~\cite{heglund1974scaling}.
\par
Another avenue we find intriguing is the role oscillators may play in a hierarchical RL setting.
Higher levels of hierarchy typically reason about task-level objectives in both a lower-dimensional space and at a slower timescale; the phase oscillators could be interpreted as a latent state with cyclic dynamics~\cite{starke2022deepphase}.
A latent state space with cyclic dynamics could serve for temporal abstraction, multi-joint coordination and amortized control for cyclic behavior~\cite{merel2019hierarchical}, a direction we find very promising.
\section*{Acknowledgements}
We thank Aditya Mehrotra for helping with robot maintenance, filming, and interesting discussions along the way. 
This work was supported by NAVER Labs.
\clearpage
\printbibliography

\end{document}